\documentclass{article}
\usepackage{ijcai21}

\usepackage{times}

\usepackage{soul}
\usepackage{url}
\usepackage[hidelinks]{hyperref}
\usepackage[utf8]{inputenc}
\usepackage[small]{caption}
\usepackage{graphicx}
\usepackage{amsmath}
\usepackage{booktabs}
\usepackage{amssymb}
\usepackage{hyperref}
\usepackage[table,xcdraw]{xcolor}
\usepackage[square,sort,comma,numbers]{natbib}
\date{November 2020}
\title{Open-World Active Learning with Stacking Ensemble for Self-Driving Cars}

\author{
Paulo R. Vieira\and
Pedro D. Félix\and
Luis Macedo
\affiliations
CISUC - Centre for Informatics and Systems of the University of Coimbra ; University of Coimbra, Portugal\\
\{pvieira, pdfelix\}@student.dei.uc.pt ;
macedo@dei.uc.pt
}

\usepackage{natbib}
\usepackage{graphicx}

\begin{document}

\maketitle

\begin{abstract}
The environments, in which autonomous cars act, are high-risky, dynamic, and full of uncertainty, demanding a continuous update of their sensory information and knowledge bases. The frequency of facing an unknown object is too high making hard the usage of Artificial Intelligence (AI) classical classification models that usually rely on the close-world assumption. This problem of classifying objects in this domain is better faced with and open-world AI approach. We propose an algorithm to identify not only all the known entities that may appear in front of the car, but also to detect and learn the classes of those unknown objects that may be rare to stand on an highway (e.g., a lost box from a truck). Our approach relies on the DOC algorithm from Lei Shu et. al. as well as on  the Query-by-Committee algorithm. 
\end{abstract}

\section{Introduction}
A large amount of problems that requires the use of AI (Artificial Intelligence) is solved with agents who know in advance all possible situations of their world. In these domains, most of the times characterised by deterministic, finite environments, the closed-world AI approach is appropriate to be employed. However, there are situations in which it is unfeasible to know in advance, at training time, the set of classes for all the objects that the agent will face in the real world, and therefore, the training phase needs to be extended to the real world, being a continuous act along the the entire life of the agent. For these situations, an open-world AI approach is more appropriate, because it allows dealing with a considerably large number of objects, whose classes are not known in advance \cite{ClosedSet,ClosedSet2,doc,NT}. This is the case of the autonomous driving domain. The amount of data needed to train the car with a considerable success in simulation environments is unimaginable large (persons, cars, buildings, trees, etc., and a huge amount of their configurations that makes decision-making a complex task). So, no matter how hard we try to get an acceptable accuracy at the training phase, when the self-driving car goes to the real world setting, there can always be one more unaccounted element like a loose tire or box in the middle of the road that is hard  to be recognised by the classification model of the autonomous vehicle.


Active learning \cite{Settles2012,Settles2011} can help solving this problem, in that, through a rinse and repeat method, active learning algorithms try first to classify the objects they select and, if still in doubt, they seek help from an oracle that guides him to the correct classification of the object.

The problem could be formally defined as follows: given a set of training data $D = \{[x_1, y_1], [x_2, y_2], ..., [x_n, y_n]\}$, where $x_i$ is the $i$-th image to train an agent and $y_i$ is the label of that image, where $y_i \in {l_1,l_2,...,l_n} = Y$, the goal is to build a classifier, $f(x)$, that can classify each image $x$ to one of the $m$ known classes in $Y$ or classify it is as unknown, i.e., not belonging to the $Y$ set. In the later case, i.e., being unknown, the goal is to ask the oracle to label the image and use this new image to the next training process. For this, it is necessary to extend the set $Y$ to $C = \{unknown, l_1, l_2, ..., l_n\}$, where \textit{n} is the number of known labels.

There are already some solutions to this problem in the literature. For instance, Deep Open Classification (DOC) \cite{doc} uses a one dimension convolutional neural network as their agent with a sigmoid output layer so that all the unknown labels have a negative impact on the probability of the outcome. Then, they evaluate the probability based on a threshold, and if it is less that the determined threshold, they reject the given object. This evaluation follows a Gaussian distribution, and the rejection occurs if the probability calculated is $\alpha$ times bigger than the standard deviation. Although it offers a good solution, this solution does not consider building a growing neural network with more information and labels because the problems did not revolve around expanding theirs labels as it is need in the self-driving cars setting.

Query-by-Committee \cite{Settles2012,Settles2011} is another promising approach to help solving this problem. It is based on the idea of a committee of agents learning different images and, as a group, find the most probable label for an image.


Since we want to create a model that learns about new classes and not reject them, i.e., turning unknown data to known, we have to embrace situations in which the highest probability is low, since these are the ones that will be dealt by the oracle. Note that this is a policy to some extent opposed to that of DOC \cite{doc}, which rejects this low highest probability instances of data. In addition, we consider the active learning strategy of Query-by-Committee (QBC) \cite{Settles2012,Settles2011}. So we implemented an oracle that, whenever we find a labelled image with the highest probability label, according to the committee, lower than a threshold $\alpha$, it is sent to the oracle so that we can give it a more accurate label. Afterwards, we use this image to train our agents as a possible new entity.

The next section describes in more detail the materials and methods we rely on, then we present the results, discuss them, and finally present our conclusions.

\section{Materials and Methods}

Our methodology comprises the following steps:

\begin{enumerate}
    \item Generate N models (Convolutional Neural Nets -- CNNs) each with a specific part of the data set and with their own specific labels;
    \item Train each of the models separately.
    \item Stack the trained models, obtaining a stacked committee, and train it with a Logistic Regression Model;
    \item Verify the output of the stacked committee and relabel entries when certainty is less than $\alpha$ with the help of the oracle.
\end{enumerate}

\subsection{Data sets and their treatment}

\begin{figure}[ht!]
\centering
\includegraphics[width=0.3\textwidth]{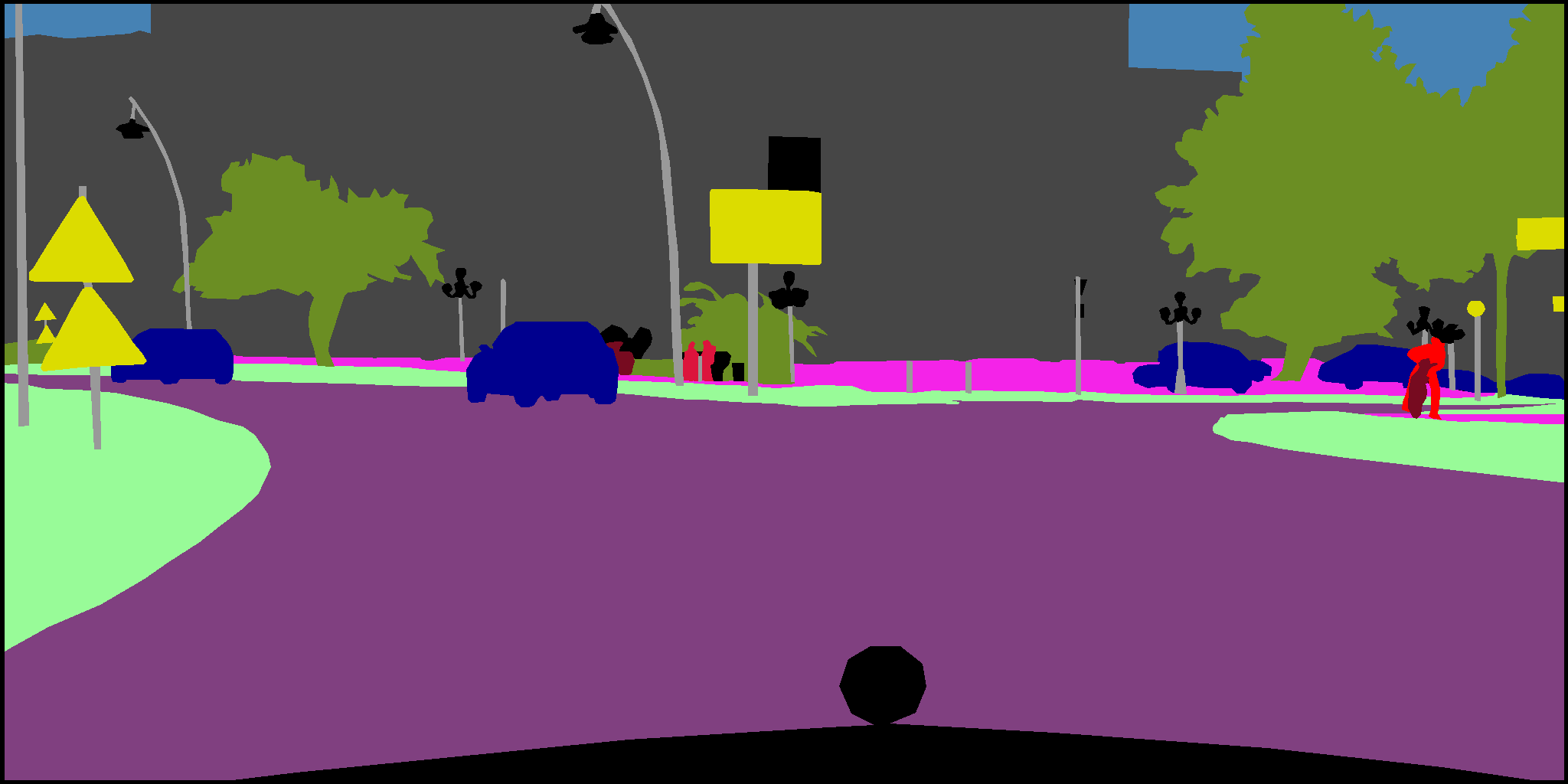}
\caption{Example of an image from the data set.}
\label{fig:datasetImage}
\end{figure}

For our study, we considered Cityscapes Datasets \cite{Cordts2016} \footnote{https://www.cityscapes-dataset.com}, which provide information focused on the single view of the driver, including images about what is in front of the car, i.e., the visual field of the driver and the location of all objects in the image (see Figure \ref{fig:datasetImage}).
These data sets were used for testing, training and validation with their different levels of complexity. We separated them into folders accordingly to the cities they belong to, each one with 10 images. We used 150 images for training and 50 for testing and re-training. These were randomly selected and, in addition, we made combinations of these data sets, choosing 15 cities for training and 5 for testing, for handling later in the construction phase of the models.

Each element of the data sets have four different components:\begin{itemize}
    \item a colored image with all the elements present in the range of view.
    \item an image which only contains both vehicles and pedestrians.
    \item a gray scaled image with all the elements present in the range of view; the more the intensity of each represented object, the  more its proximity to the car;
    \item a JSON file containing the different labels of the elements in the pictures as well as the pixel positions of the object borders.
\end{itemize}


For handling the different sized objects present in each picture, we started by transforming the frontier points of each object in the JSON file into an array vector. Then, we made a minimal area cut, which cuts the object of the image in the minimal size possible, making sure that the object is maintained fully visible in the final cut. Finally, in order to have correct sized data so that the CNN can be trained and tested with them, we resized the cut objects to a 64 by 64.

\subsection{CNN's Structure}
\begin{figure}[ht!]
\centering
\includegraphics[width=0.3\textwidth]{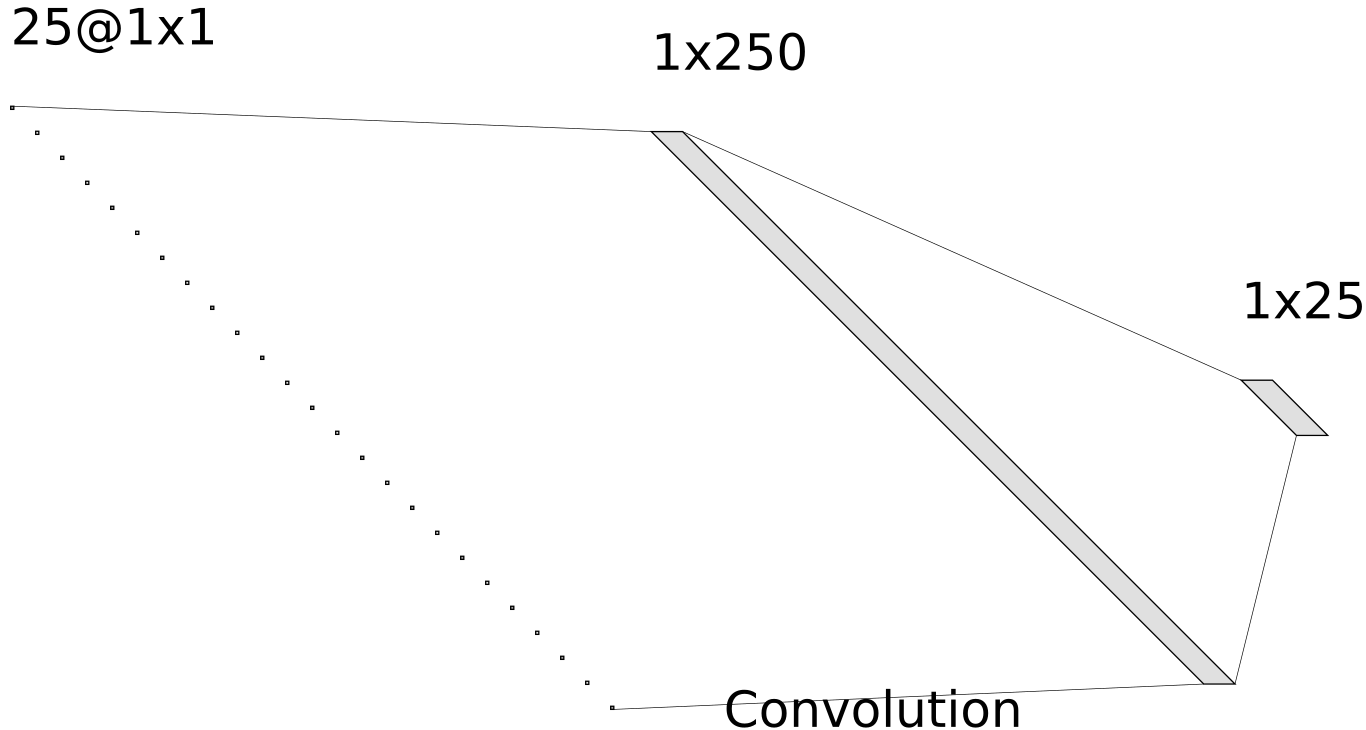}
\caption{DOC's CNN structure (the dimension of the fully connected net is 250 and there is an output of 25 classes). }
\label{fig:DOCCNN}
\end{figure}

One of the crucial points of our work is the neural networks, which have to be well designed in order to produce good results. Since we relied partially on the DOC algorithm, and acknowledging the fact that our data set comprises images, we decided to use DOC's CNN. In Figure \ref{fig:DOCCNN}, it is possible to see part of the CNN structure of DOC, which consists of an embedding layer, followed by a convolutional layer and a max pooling layer, two fully connected layers with one intermediate ReLU activation layer and finishing off with an output with the dimension of the first fully connected layer.

\begin{figure}[ht!]
\centering
\includegraphics[width=0.5\textwidth]{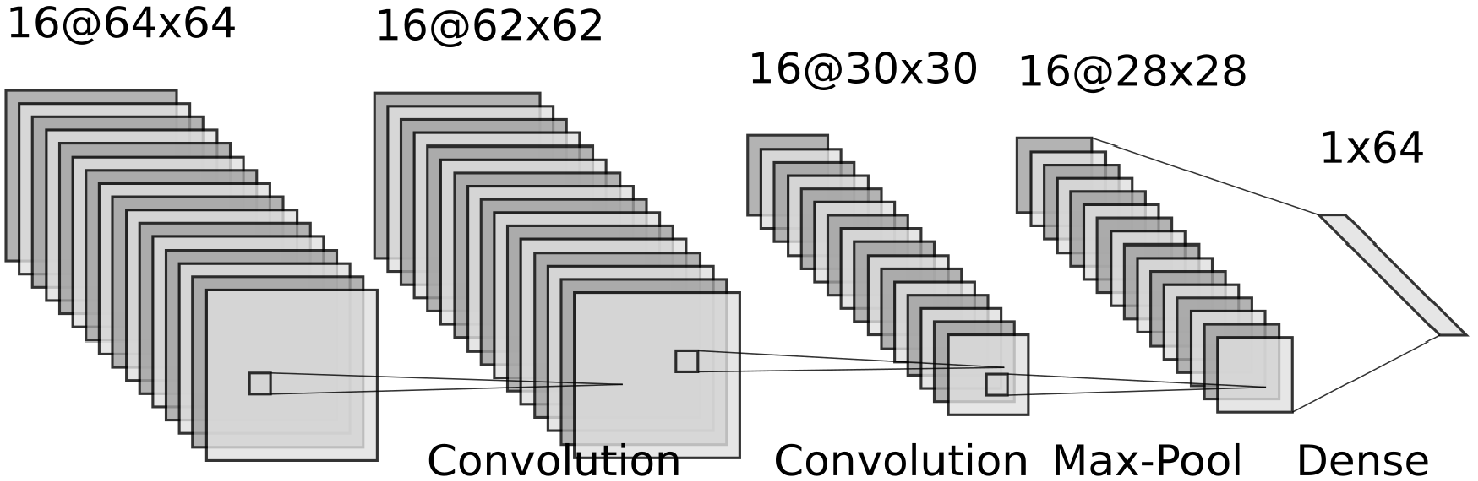}
\caption{Structure of the first (A) convolutional neural network built}
\label{CNNA}
\end{figure}

Despite being built with this general visualization of what we set to accomplish, as we are using images as inputs and DOC uses text, we could not take advantage of their first embedding layer, and so, some small changes were made, trying to achieve the highest accuracy overall. With this aim, we created three slightly different CNN's, that we will call A, B, and C. CNN A has two 2-dimensional convolutional layers and a 4 by 4 max-pooling layer after the second convolutional layer. After this, we added a fully connected layer and an ``adaptive'' output layer for dealing with N+1 labels (N for the known, one for the unknown). Figure \ref{CNNA} presents an example of this CNN. 

\begin{figure}[ht!]
\centering
\includegraphics[width=0.5\textwidth]{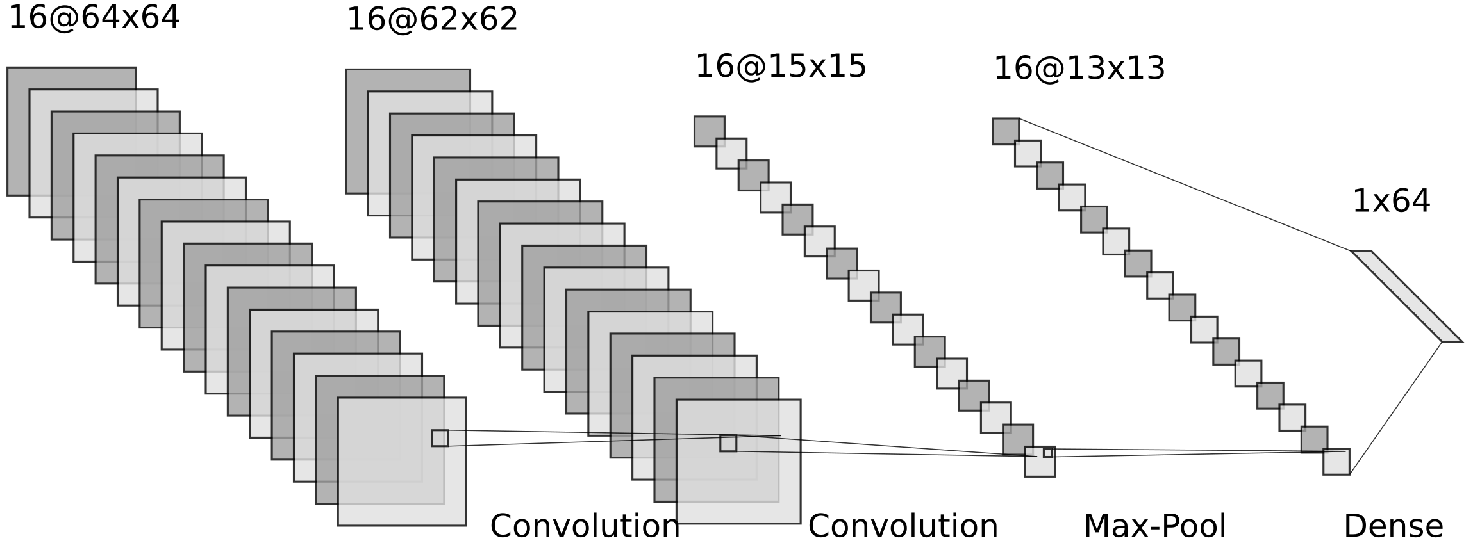}
\caption{Structure of the second (B) convolutional neural network built}
\label{CNNB}
\end{figure}

CNN B consists of the same number of convolutional layers of A, but we changed the max-pooling layer from a 4 by 4 to a 2 by 2, and we added a dropout layer that prevents overfitting by 20\%. This is followed by a fully-connected layer and the ``adaptive'' output layer, as it can be seen in Figure \ref{CNNB}.

\begin{figure}[ht!]
\centering
\includegraphics[width=0.5\textwidth]{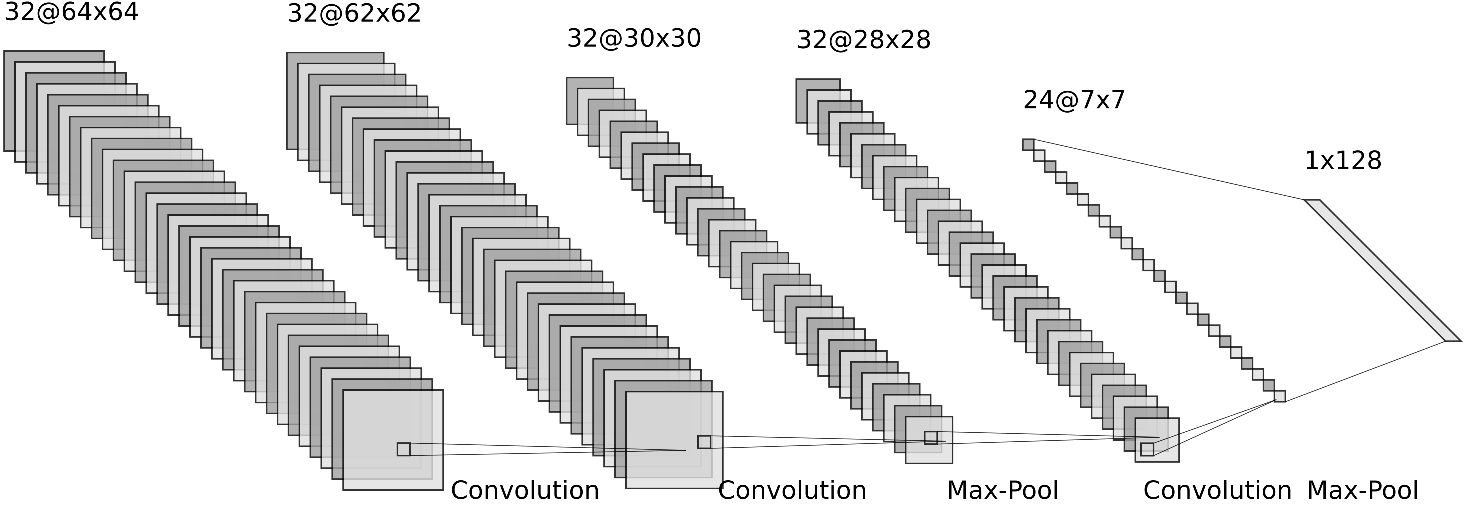}
\caption{Structure of the third (C) convolutional neural network built}
\label{CNNC}
\end{figure}

Lastly, for CNN C, we built three 2-dimensional convolutional layers with one 2 by 2 max-pooling layers after the second, and a 4 by 4 max-pooling layer after the third convolutional layer, in order to downsize the images, followed by a dropout layer that prevents overfitting by 20\%. Then, we added a fully-connected layer and finally the ``adaptiv'' output layer. Figure \ref{CNNC} presents an example of the architecture of this model.

All the CNN's were trained from scratch with the Cityscapes dataset.

\subsection{Stacking Ensemble}
With the built models described in the previous subsection, the last step is to get a model that averages the results obtained from them in order to receive the most possible information of the conjunction of all the labelled information that we have. For this, we used the simple Stacking Ensemble strategy, where all the models have the same contribution to the final output, opting by a Separate Stacking Model utilizing Logistic Regression to take in the values among the different models, selecting the best label representing the picture, with a degree of certainty, to achieve the best results. For our experiment, this works well but when taking into account a real world self driving scenario one might take a more viable approach of weighing the different models for the respective data that they contain in terms of danger level.

\subsection{Unknown Detection}
Although the Stacking Ensemble provides us with the good results, we cannot take for granted that the label provided is the correct one, since committee members may come across new objects. To this end, we use the degree of certainty to help us identify misclassified images. In the event that a new object arrives to the system, the degree of certainty will always be low when compared to known objects. Therefore, we create a minimum $\alpha$ limit that, when the degree of certainty is lower than this value, it is considered unknown.

\subsection{Experimental Setup}
\begin{figure}[ht!]
\centering
\includegraphics[width=0.5\textwidth]{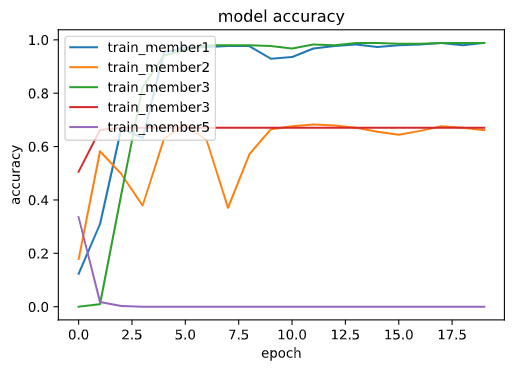}
\caption{Classification accuracy of cars and pedestrians for the different five members of a committee}
\label{accuracy}
\end{figure}

In order to fulfill our goal, we need to set some parameters so that more definite conclusions can be achieved with our results. Those parameters are: the number of epochs to train, the number of elements in the committee, the initial number of known labels and how many labels are there in the system.
We considered 6 labels, being these cars, people, vegetation, buildings, traffic signs and traffic lights. As cars and people are the most important entities while driving, since one of the primary goals of driving is to avoid hurting others or damaging the car, we take these two as a starting point in known labels. With this set, we can start to estimate the number of epochs necessary to train in order to minimize the time needed for each training. Figure \ref{accuracy} plots the accuracy of our model during 20 epochs. Considering this and several other similar results, we noticed a stabilization around 10 epochs, so we set the training phase to this number of epochs.

The difference between the members in the committee relies in the data that they had for training. The full training data set was randomly split in equivalent parts between the different members in order to best accustom them to the different aspects the elements could appear.

Last but not least, to set the size of our committee, we first needed to understand the impact of different sizes in prediction performance. With this aim, we used a committee of size 2, 3 and 5, and then we evaluated their predictions and their impact.

\section{Results}
When looking at some runs (Tables \ref{tab:my-table1}, \ref{tab:my-table2}, and \ref{tab:my-table3}) based on different data sets in a closed set of labels and with the different size of members, we can see that there is an increase of accuracy when passing to the Stacked Network, which demonstrates that there is something to gain with this approach.

\begin{table}[ht!]
\centering
\begin{tabular}{|
>{\columncolor[HTML]{BDCAF1}}c |c|c|c|}
\hline
      & \cellcolor[HTML]{BDCAF1}M\_1 & \cellcolor[HTML]{BDCAF1}M\_ 2 & \cellcolor[HTML]{BDCAF1}Stacked \\ \hline
Run 1 & 0,97                         & 0,289                         & 0,982                           \\ \hline
Run 2 & 0,261                        & 0,739                         & 0,739                           \\ \hline
Run 3 & 0,739                        & 0,713                         & 0,986                           \\ \hline
Run 4 & 0,261                        & 0,703                         & 0,972                           \\ \hline
Run 5 & 0,739                        & 0,727                         & 0,99                            \\ \hline
\end{tabular}
\caption{Accuracy of the different two members versus the Stacked Network}
\label{tab:my-table1}
\end{table}

\begin{table}[ht!]
\centering
\begin{tabular}{|
>{\columncolor[HTML]{BDCAF1}}c |c|c|c|c|}
\hline
      & \cellcolor[HTML]{BDCAF1}M\_1 & \cellcolor[HTML]{BDCAF1}M\_ 2 & \cellcolor[HTML]{BDCAF1}M\_3 & \cellcolor[HTML]{BDCAF1}Stacked \\ \hline
Run 1 & 0,962                        & 0,739                         & 0                            & 0,97                            \\ \hline
Run 2 & 0,984                        & 0                             & 0,721                        & 0,99                            \\ \hline
Run 3 & 0,727                        & 0,727                         & 0,729                        & 0,988                           \\ \hline
Run 4 & 0,994                        & 0,739                         & 0,976                        & 0,992                           \\ \hline
Run 5 & 0                            & 0,739                         & 0,974                        & 0,992                           \\ \hline
\end{tabular}
\caption{Accuracy of the different three members versus the Stacked Network}
\label{tab:my-table2}
\end{table}

\begin{table}[ht!]
\centering
\begin{tabular}{|c|c|c|c|c|c|c|}
\hline
\rowcolor[HTML]{BDCAF1} 
                              & M\_1  & M\_ 2 & M\_3  & M\_4  & M\_5  & Stacked \\ \hline
\cellcolor[HTML]{BDCAF1}Run 1 & 0,717 & 0,739 & 0,992 & 0,99  & 0,737 & 0,99    \\ \hline
\cellcolor[HTML]{BDCAF1}Run 2 & 0,739 & 0,727 & 0     & 0,739 & 0,727 & 0,982   \\ \hline
\cellcolor[HTML]{BDCAF1}Run 3 & 0,713 & 0,739 & 0,719 & 0,667 & 0,695 & 0,988   \\ \hline
\cellcolor[HTML]{BDCAF1}Run 4 & 0,683 & 0,954 & 0,255 & 0,954 & 0,261 & 0,986   \\ \hline
\cellcolor[HTML]{BDCAF1}Run 5 & 0,739 & 0,259 & 0,261 & 0,739 & 0,937 & 0,972   \\ \hline
\end{tabular}
\caption{Accuracy of the different five members versus the Stacked Network}
\label{tab:my-table3}
\end{table}

Considering these results, we opted to verify the performance when fully operating in an open set environment, working with 5 members.

\begin{figure}[ht!]
\centering
\includegraphics[width=0.5\textwidth]{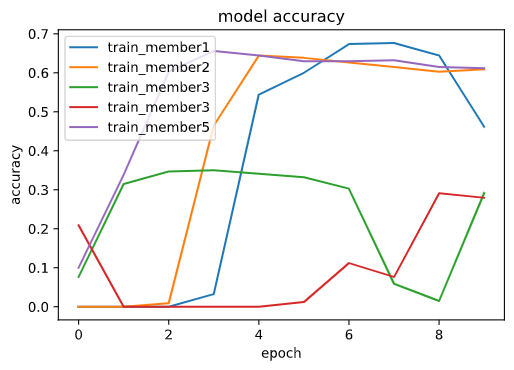}
\caption{Classification accuracy of cars and pedestrians when training our models A with 2 known labels: cars and people}
\label{trainA}
\end{figure}

\begin{figure}[ht!]
\centering
\includegraphics[width=0.5\textwidth]{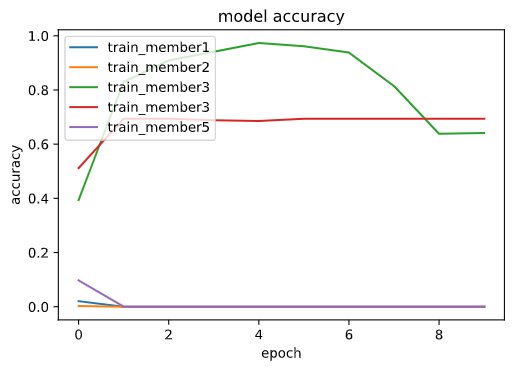}
\caption{Classification accuracy of cars and pedestrians when training our models B with 2 known labels: cars and people}
\label{trainB}
\end{figure}

\begin{figure}[ht!]
\centering
\includegraphics[width=0.5\textwidth]{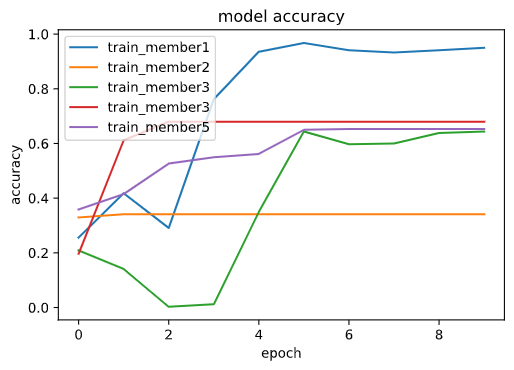}
\caption{Classification accuracy of cars and pedestrians when training our models C with 2 known labels: cars and people}
\label{trainC}
\end{figure}

Having all parameters set, we now present the results obtained when training and testing with and without unknowns. First, we needed to train our members of the committee (for the A, B and C CNN models) from scratch. We fed them only with images of people and cars, so that the model could identify the most basic and important entities that they could come across while driving. Figures \ref{trainA}, \ref{trainB} and \ref{trainC} present the accuracy obtained for this training. 

As it can seen from the models, it seems that they have very different accuracy values for only two known entities with no unknowns being presented to them. Although the accuracy of each member seems quite low, the stacked network had an accuracy of 0.99 for A, 0.976 for B and 0.94 for C. This means that they were all able to identify most of the cars and people in the given images.

But, after passing this initial training with only cars and people in mind, we put CNNs A, B and C to a further and harder test. We checked if they could still interpret the entities as well as detect unknowns. Table \ref{tab:my-table4} presents the results of this test.

\begin{figure}[ht!]
\centering
\includegraphics[width=0.5\textwidth]{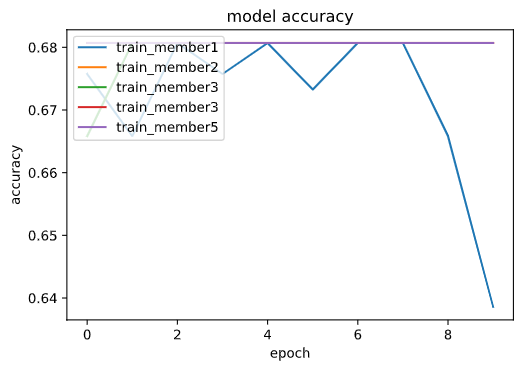}
\caption{Accuracy when training our models C with 2 known labels, 1 unknown label: traffic sign.}
\end{figure}

\begin{table}[ht!]
\centering
\begin{tabular}{|c|c|c|c|c|c|c|}
\hline
\rowcolor[HTML]{BDCAF1} 
                             & M\_1  & M\_2  & M\_3  & M\_4  & M\_5  & Stacked \\ \hline
\cellcolor[HTML]{BDCAF1}S\_A & 0,476 & 0,476 & 0,166 & 0,166 & 0,476 & 0,639   \\ \hline
\cellcolor[HTML]{BDCAF1}S\_B & 0     & 0     & 0,819 & 0,476 & 0     & 0,63    \\ \hline
\cellcolor[HTML]{BDCAF1}S\_C & 0,346 & 0,476 & 0,476 & 0,476 & 0,476 & 0,637   \\ \hline
\end{tabular}
\caption{Accuracy of the different members and Stacked Ensemble for 2 known labels, 1 unknown label for the CNN structures A, B and C.}
\label{tab:my-table4}
\end{table}

According to Table \ref{tab:my-table4}, we can see a deterioration on the accuracy of all Stacked Networks as well as in any member of any CNN structure. Despite the deterioration, they were all still able to identify around 60\% of the objects they encounter.

\begin{figure}[ht!]
\centering
\includegraphics[width=0.5\textwidth]{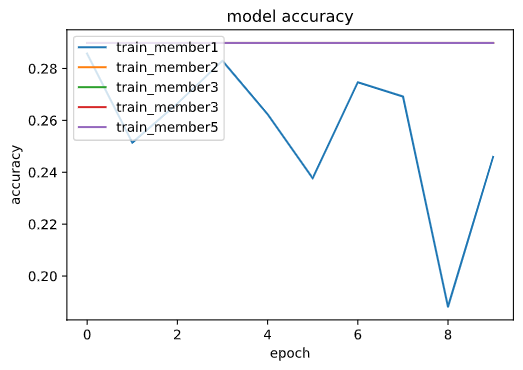}
\caption{Accuracy when training our models C with 3 known labels, 1 unknown label: traffic light}
\end{figure}

\begin{table}[ht!]
\centering
\begin{tabular}{|c|c|c|c|c|c|c|}
\hline
\rowcolor[HTML]{BDCAF1} 
                             & M\_1  & M\_2  & M\_3  & M\_4  & M\_5  & Stacked \\ \hline
\cellcolor[HTML]{BDCAF1}S\_A & 0,409 & 0,409 & 0,144 & 0     & 0,409 & 0,546   \\ \hline
\cellcolor[HTML]{BDCAF1}S\_B & 0     & 0     & 0,819 & 0,476 & 0     & 0,63    \\ \hline
\cellcolor[HTML]{BDCAF1}S\_C & 0,293 & 0,409 & 0,409 & 0,409 & 0,409 & 0,454   \\ \hline
\end{tabular}
\caption{Accuracy of the different members and Stacked Ensemble for 3 known labels, 1 unknown label for the CNN structures A, B and C}
\label{tab:my-table5}
\end{table}

At that stage, our models knew three new entities, and it was time to add a forth entity, unknown to all of them. We were still able to notice a decline in the accuracy with the increment of new objects in the images, although not being as abrupt as for the first unknown. It is also possible to see a small decrease on the accuracy of the members of the committee. 

We then presented a third unknown variable to our committees and expect a new decrease on the accuracy overall.

\begin{figure}[ht!]
\centering
\includegraphics[width=0.5\textwidth]{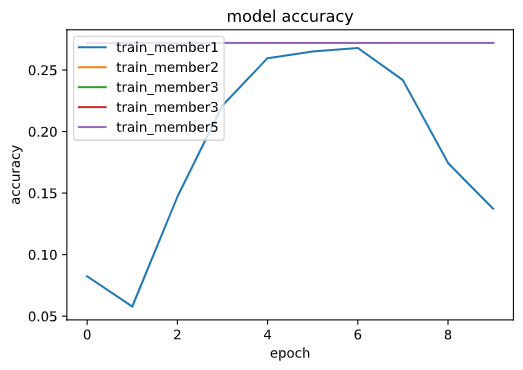}
\caption{Accuracy when training our models C with 4 known labels, 1 unknown label: building}
\label{fig:trainC4k1u_building}
\end{figure}

\begin{table}[ht!]
\centering
\begin{tabular}{|c|c|c|c|c|c|c|}
\hline
\rowcolor[HTML]{BDCAF1} 
                             & M\_1  & M\_2  & M\_3  & M\_4  & M\_5  & Stacked \\ \hline
\cellcolor[HTML]{BDCAF1}S\_A & 0,121 & 0,122 & 0     & 0,189 & 0,121 & 0,087   \\ \hline
\cellcolor[HTML]{BDCAF1}S\_B & 0     & 0     & 0,54  & 0,409 & 0     & 0,543   \\ \hline
\cellcolor[HTML]{BDCAF1}S\_C & 0,279 & 0,409 & 0,409 & 0,409 & 0,409 & 0,446   \\ \hline
\end{tabular}
\caption{Accuracy of the different members and Stacked Ensemble for 4 known labels, 1 unknown label for the CNN structures A, B and C.}
\label{tab:my-table6}
\end{table}

Again, we are able to see, now in Table \ref{tab:my-table6}, that the unknown objects worsen our accuracy to detect our objects. We can even notice an abrupt decay on the stacked network of architecture A and for the first time have the stacked network with a lower accuracy than most of the members of the respective committee.

\begin{figure}[ht!]
\centering
\includegraphics[width=0.5\textwidth]{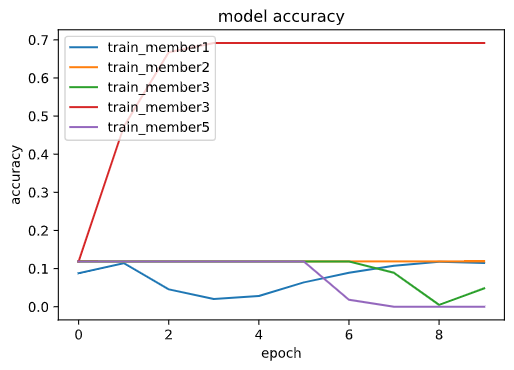}
\caption{Accuracy when training our models C with 5 known labels, 1 unknown label: vegetation}
\label{fig:trainC5k1uVeg}
\end{figure}

Figure \ref{fig:trainC5k1uVeg} reports the results of the addition of the last unknown entity: vegetation. As this is a complex element, we confirmed the expectations that the results get worse.

\begin{table}[ht!]
\centering
\begin{tabular}{|c|c|c|c|c|c|c|}
\hline
\rowcolor[HTML]{BDCAF1} 
                             & M\_1  & M\_2  & M\_3  & M\_4  & M\_5  & Stacked \\ \hline
\cellcolor[HTML]{BDCAF1}S\_A & 0,121 & 0,122 & 0     & 0,189 & 0,121 & 0,087   \\ \hline
\cellcolor[HTML]{BDCAF1}S\_B & 0     & 0     & 0,163 & 0,163 & 0     & 0,051   \\ \hline
\cellcolor[HTML]{BDCAF1}S\_C & 0,121 & 0,288 & 0,288 & 0,288 & 0,288 & 0,263   \\ \hline
\end{tabular}
\caption{Accuracy of the different members and Stacked Ensemble for 5 known labels, 1 unknown label for the CNN structures A, B and C}
\label{tab:my-table7}
\end{table}

What we stated for the test, stands for this last unseen element but for different structures: now, for architectures B and C, with more impact on B, due to the complexity of this object as stated above.

\section{Discussion and Conclusions}

Overall, we may conclude that due to a back and forward in the construction of the best architecture, we got a structure that gave us good results even though they were not optimal. 


A problem that we faced, not fully related with the networks structures, was the lack of related works, which would be valuable for comparison. So in order to make comparisons, we had to be confined to the different networks we created in order to get a final one.

When handling initial versions of our models, when making attempts with the introduction of four or more unknown labels to the network, the models abruptly passed from accuracies (in the Stacked model) above 40\% to values with single digits between 2\% and 8\% (A and B). But, as we can see in the final iteration of our model (C), when dealing with the fifth unknown label we maintain an accuracy over 20\% (this difference between models can be seen in Table \ref{tab:my-table7}).

As it was already mentioned when talking about Stacking Ensemble, in the future, in order to reach better performances, we plan to chose a different variant of Stacking Ensemble, in which, rather than being ruled by the Separate Stacking Model, which only makes use of Logistic Regression to get the best results of the agglomerate of models, we could try and make use of an Integrated Stacking Model, which inserts the several models in the committee and embeds them in a larger multi-headed neural network that could be trained to better combine the different predictions.
    
A final struggle that we were also faced with was the complexity that some elements had in the data images. In some cases, even to the human eye, it was hard to come up with a class of the target object in the image, due to obstructions of other objects in between the element and the oracle or committee/models that was analyzing it. And this could also be a factor for some unbalanced training and testing, even though we tried to avoid it, when separating the different cities into folders, to make a selection which limited the amount of covered objects.


\end{document}